\documentclass[11pt]{article}
\usepackage{amssymb}

\usepackage[preprint]{acl}

\usepackage{times}
\usepackage{latexsym}

\usepackage[T1]{fontenc}
\usepackage[utf8]{inputenc}

\usepackage{microtype}
\usepackage{colortbl}
\usepackage{booktabs}
\usepackage{multirow}
\usepackage{graphicx}
\usepackage[table,xcdraw]{xcolor}
\usepackage{geometry}
\definecolor{highlightrow}{gray}{0.95} % 你的模型行的背景色
\definecolor{avggray}{gray}{0.9}     % Avg 列的背景色\usepackage{inconsolata}
\definecolor{oursbg}{gray}{0.95} % 定义高亮背景色

\definecolor{c_good}{HTML}{588E31} % <10%
\definecolor{c_mid}{HTML}{2E54A1}  % 10-20%
\definecolor{c_bad}{HTML}{C81D31}  % >20%
\definecolor{oursbg}{RGB}{235, 242, 250}
\usepackage{booktabs}
\usepackage{amsmath}
\usepackage{tcolorbox}
\usepackage{enumitem}
\usepackage{times}
\usepackage{latexsym}
\usepackage{booktabs}
\usepackage{amssymb}
\usepackage{pifont}    % 用于 \ding{51} (打钩) and \ding{55} (打叉)=
\usepackage{xcolor}    % 用于定义颜色
\usepackage[ruled,vlined]{algorithm2e}
\usepackage{caption}

\usepackage[utf8]{inputenc}
\usepackage{xcolor}
\usepackage{tcolorbox}
\tcbuselibrary{skins, breakable, listings} % 必须加载这些库
\usepackage{enumitem}

\definecolor{promptbg}{RGB}{245,247,250} % 极淡的蓝灰色背景
\definecolor{promptframe}{RGB}{52,73,94} % 深蓝灰色边框
\definecolor{jsonkey}{RGB}{128,0,0}      % JSON Key颜色
\definecolor{jsonstring}{RGB}{0,128,0}   % JSON String颜色

\title{EHRWorld: A Patient-Centric Medical World Model for\\Long-Horizon Clinical Trajectories}

\author{
    \textbf{Linjie Mu\textsuperscript{1}}, 
    \textbf{Zhongzhen Huang\textsuperscript{1}}, 
    \textbf{Yannian Gu\textsuperscript{1}}, 
    \textbf{Shengqian Qin\textsuperscript{1}}, \\
    \textbf{Shaoting Zhang\textsuperscript{1,}}\thanks{\ \ Corresponding authors.}, 
    \textbf{Xiaofan Zhang\textsuperscript{1,2,}}\footnotemark[1] \\
    \textsuperscript{1}Shanghai Jiao Tong University, 
    \textsuperscript{2}Shanghai Innovation Institute
}
\begin{document}
\maketitle

\begin{abstract}
World models offer a principled framework for simulating future states under interventions, but realizing such models in complex, high-stakes domains like medicine remains challenging. Recent large language models (LLMs) have achieved strong performance on static medical reasoning tasks, raising the question of whether they can function as dynamic medical world models capable of simulating disease progression and treatment outcomes over time. In this work, we show that LLMs only incorporating medical knowledge struggle to maintain consistent patient states under sequential interventions, leading to error accumulation in long-horizon clinical simulation. To address this limitation, we introduce EHRWorld, a patient-centric medical world model trained under a causal sequential paradigm, together with EHRWorld-110K, a large-scale longitudinal clinical dataset derived from real-world electronic health records. Extensive evaluations demonstrate that EHRWorld significantly outperforms naive LLM-based baselines, achieving more stable long-horizon simulation, improved modeling of clinically sensitive events, and favorable reasoning efficiency, highlighting the necessity of training on causally grounded, temporally evolving clinical data for reliable and robust medical world modeling.
\end{abstract}

\section{Introduction}
\label{sec:intro}

\begin{figure}[t!]
    \centering
    \includegraphics[width=0.49\textwidth]{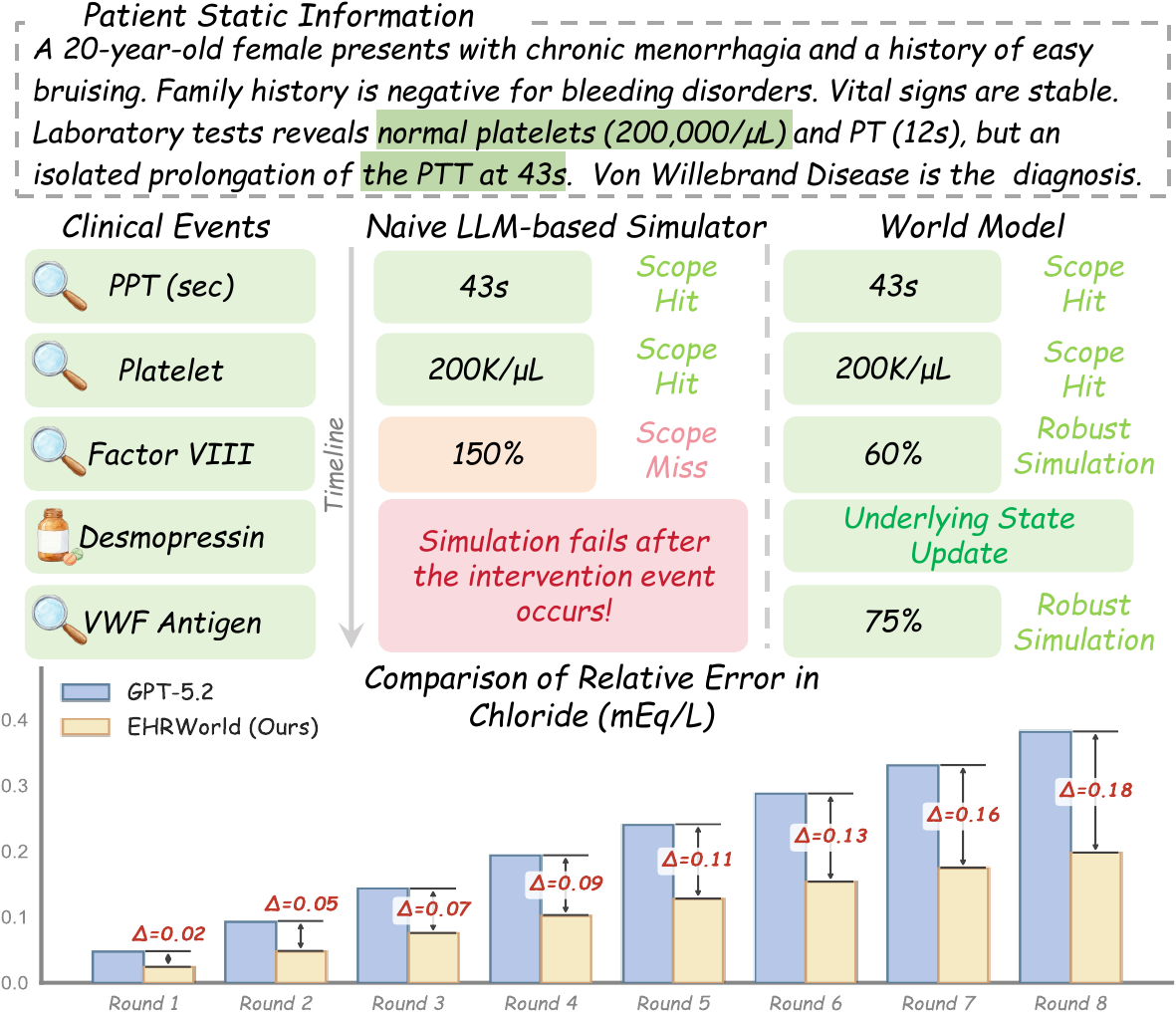}
    % \caption{
    % Comparison between naive LLM-based simulation and the proposed world model.
    % While the naive simulator matches observed values, it fails to robustly simulate interventions and exhibits error accumulation over time.
    % In contrast, the world model maintains consistent latent states, enabling reliable intervention simulation and reduced long-horizon error drift.
    % }
    % \caption{Overview of patient simulation and error accumulation. The upper part shows the patient's static clinical information. The middle part compares the simulation mechanisms of the naive LLM-based simulator and the world model, highlighting the naive method's failure to handle Scope Miss issues and its inability to simulate interventions. The lower part presents the error accumulation plot, where the naive simulator exhibits increasing error, while the world model maintains stability with reduced error drift, demonstrating superior robustness in modeling long-horizon simulations.}
    \caption{{Overview of patient simulation challenges and performance evaluation.} The upper and middle panels illustrate a clinical scenario where a standard LLM-based simulator fails to infer implicit physiological states or correctly update patient status following medical interventions. In contrast, the proposed EHRWorld model maintains logical consistency and robustness. The lower panel presents the trajectory of relative error for Chloride levels across eight simulation rounds. We compare our approach against GPT-5.2, demonstrating that our model significantly constrains the rate of error propagation, resulting in a widening performance gap that highlights robustness in long-horizon simulations.}

    \label{fig:intro}
\end{figure}

The concept of world models has emerged as a pivotal paradigm in the pursuit of general artificial intelligence~\cite{ha2018world,lecun2022path}. By constructing a comprehensive internal representation of the environment, a world model enables simulating future states conditioned on previous actions~\cite{hafner2019dream}. This capability supports \textit{planning in imagination} prior to real-world execution~\cite{bengio2019system}. Such an ability to reason about how current interactions shape future dynamics is fundamental to effective autonomous decision-making in complex and evolving environments, yet remains challenging to realize in real-world, high-stakes domains.

As a multifaceted endeavor, medicine provides a particularly compelling setting for world models. Clinical practice involves highly complex systems in which heterogeneous signals, ranging from physiological measurements and imaging to laboratory tests and clinical narratives, interact over time. Clinical decisions, such as medication choice, dosage, and timing, can alter a patient’s future physiological trajectory, often in ways that are uncertain and patient-specific. Consequently, clinicians must routinely perform counterfactual reasoning, implicitly asking how outcomes might differ under alternative treatment plans. Accurately modeling long-horizon patient state transitions under various treatment strategies is therefore essential for effective and personalized care. A medical world model could help clinicians ``see'' health as a continuous evolving process, while also providing a principled foundation for AI systems that estimate and leverage the causal effects of clinical actions.

Recent advances in large language models (LLMs) have significantly reshaped the landscape of medical AI. These models have demonstrated strong performance across a range of medical tasks, including clinical report generation~\cite{wang2023r2gengpt,jin2024promptmrg}, diagnostic reasoning~\cite{dou2025baichuan,chen2024huatuogpt}, and medical question answering~\cite{li2023llava,mu2025medceg}. Motivated by these successes, a natural question arises: \textit{can LLMs, trained on vast corpora of medical textbooks, literature and clinical case reports, serve as world models capable of simulating the temporal evolution of clinical indicators and treatment outcomes? }

% We examine whether LLMs trained on static medical corpora can function as medical world models by analyzing their behavior in sequential clinical simulation. While such models can reproduce observed clinical facts at individual time steps, they often struggle to maintain consistent patient states when simulating interventions over time. As illustrated in Figure~\ref{fig:intro}, this results in scope miss and accumulating errors across multi-step interactions, reflecting the absence of explicit mechanisms for tracking latent physiological states. Together, these observations suggest that strong static reasoning alone is insufficient for reliable, intervention-aware clinical simulation.

% We examine whether LLMs trained on static medical corpora can function as medical world models by analyzing their behavior in sequential clinical simulation. While such models can reproduce observed clinical facts at individual time steps, they often struggle to internally maintain consistent patient states when simulating interventions over time. As illustrated in Figure~\ref{fig:intro}, this results in scope miss and accumulating errors across multi-step interactions, reflecting the absence of explicit mechanisms for tracking underlying latent physiological states. Together, these observations suggest that strong static reasoning alone is insufficient for reliable, intervention-aware clinical simulation.

We examine whether LLMs incorporating medical knowledge can function as medical world models by analyzing their behavior in sequential clinical simulations. As illustrated in Figure~\ref{fig:intro}, while these models can accurately replicate clinical observations at individual time points, they struggle when the simulation moves beyond static information. Meanwhile, they struggle to internally maintain consistent patient states when intervention events occur. These phenomena result in the accumulation of errors across multi-step interactions, reflecting the absence of explicit mechanisms for tracking underlying physiological states.

To address these limitations, we establish a robust data foundation by curating a large-scale clinical dataset, EHRWorld-110K, derived from real-world Electronic Health Records (EHRs)~\cite{johnson2023mimic}. The construction pipeline consists of three main stages: (1) extracting episode-level static patient profiles from unstructured clinical notes; (2) organizing temporally ordered event sequences from time-stamped event logs; and (3) integrating patient profiles with event sequences at the hospitalization-episode level, followed by rigorous quality filtering. In total, EHRWorld-110K comprises approximately 110 thousand diverse hospitalization episodes and 17.5 million highly clinical events, covering the full trajectory from admission to discharge. This dataset provides a principled foundation for learning patient state evolution and intervention-conditioned transitions in longitudinal real-world clinical care.

% Building on this foundation, we introduce a generative training paradigm that models clinical care as a continuous sequential process, enabling the learning of disease progression dynamics and intervention-driven physiological transitions. Under this paradigm, we train a family of  EHRWorld models at different parameter scales. Unlike prior approaches that rely on static templates, these models function as evolving patient simulators, dynamically updating latent physiological states based on interaction history and therapeutic inputs. Through comprehensive evaluations, we show that  EHRWorld substantially outperforms naive LLM-based baselines, exhibiting markedly reduced error accumulation in long-horizon simulations, improved stability on clinically sensitive events, and favorable reasoning efficiency compared to other models. These results further demonstrate the necessity of training on causally grounded, temporally evolving clinical data for reliable medical world modeling.

Building upon this foundation, we introduce a generative training paradigm that models clinical trajectories as a continuous sequential process, facilitating the learning of intervention-driven physiological transitions.
We then train a family of models, EHRWorld, at different parameter scales. EHRWorlds function as evolving patient simulators, dynamically learning and updating physiological states based on interaction history and therapeutic inputs.
Through extensive evaluations, we demonstrate that EHRWorld significantly outperforms naive LLM-based baselines, exhibiting a marked reduction in error accumulation in long-horizon simulations, enhanced stability during clinically sensitive events, and improved reasoning efficiency compared to other models.
These findings underscore the importance of training on causally grounded, temporally evolving clinical trajectory data to ensure reliable modeling.

% \vspace{0.5em}
In summary, our main contributions are:
\begin{itemize}
    \item We introduce EHRWorld-110K, a large-scale longitudinal dataset that captures complete high-fidelity patient clinical care trajectories from admission to discharge, enabling the study of temporally evolving and underlying intervention-conditioned clinical dynamics.
    
    \item We propose EHRWorld, a unified family of patient-centric medical world models trained under a causal sequential paradigm, which simulate dynamic disease progression by maintaining and updating physiological states in response to clinical interventions.
    
    \item We present a comprehensive evaluation. The results demonstrate that EHRWorld significantly outperforms naive LLM-based baselines in long-horizon clinical simulation, with reduced error accumulation, improved stability on clinically sensitive events, and favorable reasoning efficiency.
\end{itemize}

\section{Related Work}

\subsection{Naive LLM-based Simulation}

Recent medical AI evaluation has increasingly shifted from static knowledge assessment toward interactive patient simulation that better reflects real-world clinical practice. Several frameworks employ LLMs as virtual patients to enable  multi-turn clinical interaction. Representative systems such as AgentClinic~\cite{schmidgall2024agentclinic} and AutoMedic~\cite{oh2025automedicautomatedevaluationframework} use prompt-based role-playing to simulate patients with predefined clinical profiles, facilitating the evaluation of history-taking and diagnostic reasoning. Extensions such as CP-Env~\cite{zhu2025cpenvevaluatinglargelanguage} and MAQuE~\cite{gong2025dialogue} further adopt dialogue-driven settings, where medical agents iteratively query patient agents to resolve underlying and inherent clinical uncertainty.

Despite their interactive interfaces, these approaches rely on fundamentally static simulation mechanisms. Patient states are typically grounded in fixed case descriptions, resulting in immutable representations without temporal or physiological dynamics. As a consequence, such simulators are prone to hallucination when queried beyond explicitly provided information, and they do not support causal state transitions under clinical interventions. These limitations prevent existing  simulators from modeling disease progression or treatment effects over time, highlighting the need for dynamic, causally grounded medical world models.

\subsection{EHR World Models}

World models are fundamentally characterized by their capacity to internally represent a complex environment and simulate future states conditioned on agent actions~\cite{ha2018world}. In the context of EHRs, prior research has progressed from passive patient representation toward limited forms of generative modeling; however, these approaches have largely fallen short of providing truly interactive, intervention-aware simulation.

Early efforts in EHR modeling primarily focused on discriminative prediction. Models such as BEHRT~\cite{li2020behrt} and Med-BERT~\cite{rasmy2021med} leverage transformer ~\cite{vaswani2017attention} architectures to encode longitudinal patient histories for downstream risk estimation, including mortality and readmission. While effective for risk stratification, these approaches operate as passive observers: they predict outcomes from fixed sequences without modeling how patient states evolve in response to clinical interventions.

Subsequent work explored generative modeling to synthesize patient trajectories. Rule-based simulators such as Synthea~\cite{chen2019validity} and data-driven approaches like PatientSim~\cite{kyung2025patientsim} generate synthetic EHR sequences for data augmentation and benchmarking. However, these methods typically produce static trajectories and do not support dynamic state transitions conditioned on therapeutic actions, limiting their use for complex treatment planning or counterfactual reasoning. In contrast, EHRWorld models patient state evolution as a fully causal and dynamic process, enabling continuous simulation of physiological responses under sequential clinical interventions.

\section{Data Construction Pipeline}
\label{sec:data}

\begin{figure*}[t]
    \centering
    \includegraphics[width=0.98\textwidth]{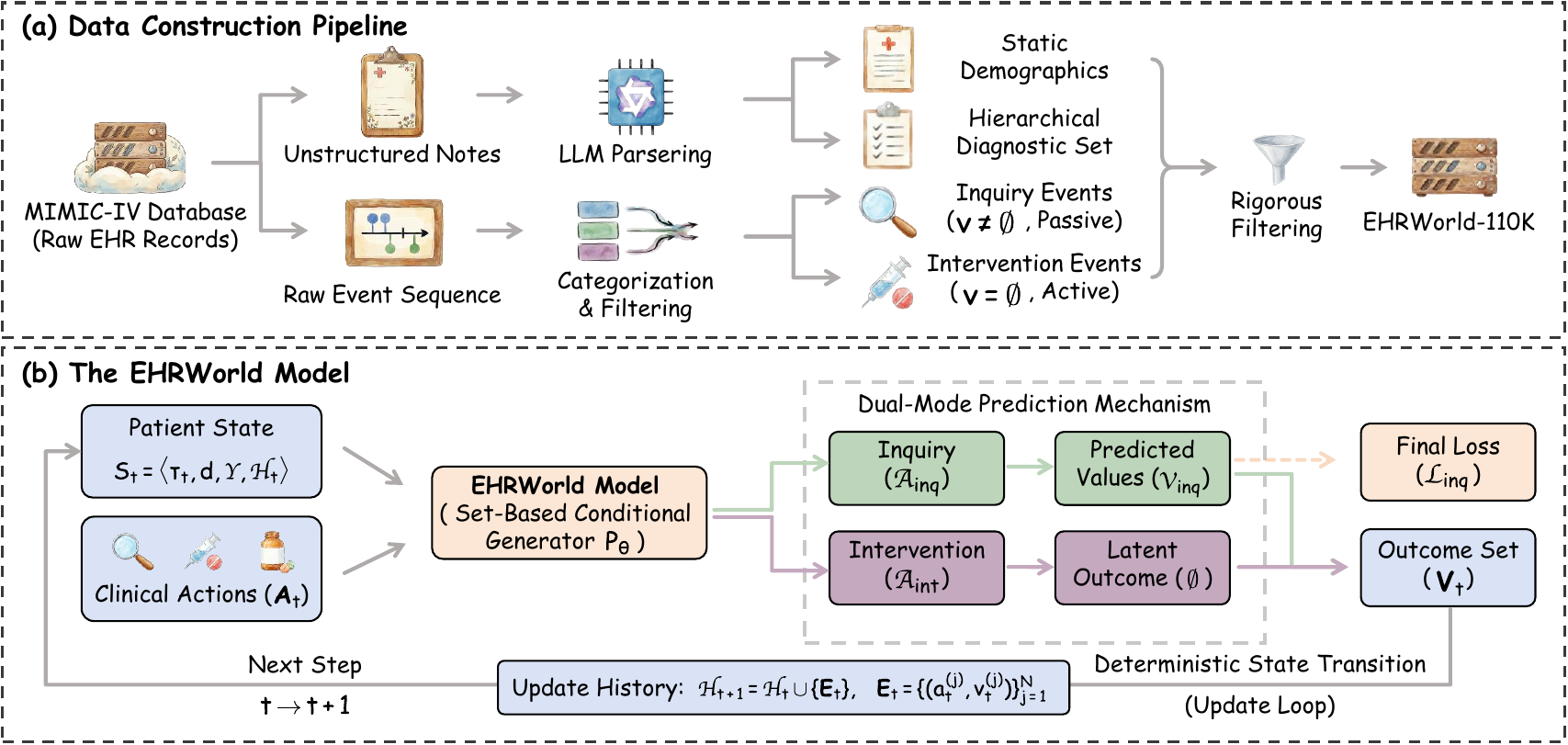}
    \caption{
Overview of the EHRWorld framework.
(a) Data construction pipeline.
Raw EHR records from MIMIC-IV are processed into the EHRWorld-110K dataset by integrating static patient context with longitudinal clinical events.
(b) The EHRWorld model.
At each step, the model conditions on the current patient state and a set of clinical actions, generates outcomes via a dual-mode mechanism for inquiries and interventions, and updates the interaction history for sequential trajectory simulation.
}
\label{fig:method}
\end{figure*}

To facilitate longitudinal and intervention-aware clinical trajectory simulation, we curate a large-scale dataset derived from real-world EHR records. In this section, we will detail the formalization of the EHRWorld-110K pipeline, as schematically illustrated in the upper panel of Figure~\ref{fig:method}.

\subsection{Data Sources and Preprocessing}

Our pipeline starts from raw records in the MIMIC-IV database~\cite{johnson2023mimic}, including unstructured clinical notes and structured event logs. These two data streams are processed in parallel and subsequently integrated at the episode level.

\paragraph{Parsing Unstructured Clinical Notes.} \mbox{}
Unstructured discharge summaries are processed using LLMs, such as \emph{Qwen3-235B-A22B-Instruct}~\cite{qwen3}, to extract patient-level static information. This step yields structured demographic attributes (e.g., age and gender) as well as a hierarchical diagnostic set, including primary and secondary diagnoses. These elements provide a stable clinical context for each hospitalization episode.

\paragraph{Processing Raw Event Sequences.} \mbox{}
In parallel, raw time-stamped clinical events are extracted and organized into event sequences spanning the entire hospital stay. Each event is categorized based on its clinical role and whether it produces an observable value. Specifically, we distinguish between:
% \begin{itemize}[leftmargin=*]
%     \item \textbf{Inquiry Events}, which correspond to passive observations of the patient state (e.g., laboratory tests and physical examinations) and yield explicit measurement values ($v \neq \emptyset$);
%     \item \textbf{Intervention Events}, which correspond to active clinical actions, such as medication administrations and procedures, that aim to modify the patient state without directly generating observable values ($v = \emptyset$).
% \end{itemize}

\begin{itemize}[leftmargin=*, noitemsep]
    \item \textbf{Inquiry Events}, which correspond to passive observations of the patient state, such as laboratory tests and physical examinations, and yield explicit measurement values;
    \item \textbf{Intervention Events}, which correspond to active clinical actions aimed at altering the patient's condition, such as medication administrations and procedures, that aim to modify the patient state without directly generating observable outcomes.
\end{itemize}

Only events that were explicitly executed are retained, while incomplete, duplicated, or administratively recorded entries are removed.

\subsection{Data Filtering and Partition}

% For each hospitalization, the parsed static information and the processed event sequence are seamlessly aggregated into a single episode, where all events are strictly ordered chronologically to faithfully preserve the temporal structure of the clinical trajectory. Building upon this structured representation, we apply rigorous filtering criteria to ensure data quality and consistency across the constructed episodes. Admissions with insufficient clinical activity or incomplete records are excluded, resulting in a finalized dataset comprising $110$ thousand hospitalization episodes with approximately $17.5$ million clinical events. These episodes constitute the  EHRWorld-110K dataset, providing high-quality longitudinal clinical trajectories suitable for downstream simulation tasks. A detailed discussion of this section is provided in Appendix~\ref{app:data_details}.
For each hospitalization, the parsed static information, including patient demographics and hierarchical diagnosis,  and event sequence are seamlessly combined into a single episode, with events strictly ordered chronologically to preserve the temporal structure of the clinical trajectory. 
We then apply stringent filtering criteria to ensure data quality and consistency across the constructed episodes, excluding admissions with insufficient clinical activity or incomplete records. 
The resulting dataset, EHRWorld-110K, consists of $110,513$ hospitalization episodes and approximately $17.5$ million clinical events, offering high-quality longitudinal clinical trajectories for downstream simulation tasks. A detailed discussion is provided in Appendix~\ref{app:data_details}.

% To support reliable evaluation while preserving the diversity of real-world clinical data, we partition the constructed dataset using stratified sampling based on primary diagnostic categories. This procedure yields a held-out test set of $579$ hospitalization episodes, comprising $84{,}010$ inquiry events and $25{,}798$ intervention events. This balanced distribution ensures rigorous testing across diverse clinical conditions. The remaining episodes are used for model training.

\noindent\textbf{Dataset Partition.}\mbox{}
To support reliable evaluation while preserving the diversity of real-world clinical data, we partition the constructed dataset using stratified sampling based on primary diagnostic categories. 
This procedure yields a held-out test set of $579$ hospitalization episodes, comprising $84{,}010$ inquiry events and $25{,}798$ intervention events, and includes $1{,}043$ unique primary and secondary diagnostic conditions, ensuring a comprehensive representation of various clinical scenarios.
This balanced distribution ensures rigorous testing across diverse clinical conditions. The remaining episodes are used for model training.

% \paragraph{Expert Verification.}
% Given the structural transformations inherent in our pipeline, we engaged three board-certified physicians to audit the test set. The validation focused on two critical dimensions: \textit{Extraction Fidelity}, to ensure static profiles accurately preserve information from the original summaries; and \textit{Temporal Coherence}, to confirm the logical causality of the aligned event sequences. The audit yielded a sequence-level accuracy exceeding 98\%, validating that our pipeline maintains the ground-truth fidelity required for high-precision simulation.

\section{The EHRWorld Model}
\label{sec:model}

We formulate patient simulation as a sequential decision process in which clinical interactions evolve over discrete simulation steps indexed by $t$. Each step corresponds to a physiological timestamp $\tau_t$, and the simulator models how patient states evolve in response to sets of concurrent clinical actions. As illustrated in the lower panel of Figure~\ref{fig:method},  EHRWorld is designed as a conditional world model that explicitly tracks patient states and updates them under sequential inquiries and interventions.

\subsection{Patient State Representation}

Let $\mathcal{S}$ denote the patient state space. At step $t$, the patient state $S_t \in \mathcal{S}$ is defined as
\begin{equation}
    S_t = \langle \tau_t, d, \mathcal{Y}, \mathcal{H}_t \rangle,
\end{equation}
where $\tau_t$ denotes the current physiological timestamp, $d$ represents static demographic attributes, and $\mathcal{Y}$ denotes the diagnostic profile of the current hospitalization episode, extracted from discharge summaries and used as episode-level clinical context. To provide causal grounding, $\mathcal{Y}$ includes both primary and secondary diagnoses together with their clinical rationales:
\begin{equation}
    \mathcal{Y} = \{\langle y_{\text{pri}}, r_{\text{pri}} \rangle\} \cup \{\langle y_{\text{sec}}^{(k)}, r_{\text{sec}}^{(k)} \rangle\}_{k=1}^{K}.
\end{equation}
The interaction history $\mathcal{H}_t = \{\mathbf{E}_1, \dots, \mathbf{E}_{t-1}\}$ aggregates all past clinical events prior to $\tau_t$ and is initialized as empty, where each $\mathbf{E}_k$ contains all actions and outcomes occurring at timestamp $\tau_k$.

\subsection{Set-Based Conditional Generation}

% Clinical practice often involves issuing multiple orders simultaneously. To reflect this,  EHRWorld models clinical interaction as a set-based conditional generation problem. At step $t$, a clinician issues a set of orders, which we model as actions.
% \begin{equation}
%     \mathbf{A}_t = \{a_t^{(1)}, \dots, a_t^{(N)}\},
% \end{equation}
% % The model is required to generate a corresponding set of outcomes of xx
% The model is required to generate a corresponding set of outcomes,
% \begin{equation}
%     \mathbf{V}_t = \{v_t^{(1)}, \dots, v_t^{(N)}\},
% \end{equation}
% representing the effects of these clinical actions on the patient's state. These outcomes are necessary to capture the dynamic and concurrent nature of clinical decision-making, where multiple actions can influence patient states in parallel.

Real-world clinical practice often involves issuing multiple medical orders simultaneously. To reflect this complexity,  EHRWorld models clinical interaction as a set-based conditional generation problem. 
At step $t$, a clinician issues a set of orders, which we formally model as actions $\mathbf{A}_t = \{a_t^{(1)}, \dots, a_t^{(N)}\}$. 
The model is required to generate a corresponding set of outcomes $\mathbf{V}_t = \{v_t^{(1)}, \dots, v_t^{(N)}\}$, representing the effects of these clinical actions on the patient's state. 
These outcomes are necessary to capture the dynamic and concurrent nature of clinical decision-making, where multiple actions can influence patient states in parallel.

 EHRWorld parameterizes a conditional distribution $P_\theta$ over outcome sets given the current state and action set:
\begin{equation}
    P_\theta(\mathbf{V}_t \mid S_t, \mathbf{A}_t) = \prod_{j=1}^{N} P_\theta(v_t^{(j)} \mid S_t, \mathbf{A}_t),
\end{equation}
allowing the model to generate coherent outcomes for concurrent clinical actions while conditioning on a shared latent patient state.

\subsection{Dual-Mode Prediction Mechanism}

Clinical actions differ in whether they produce immediate observable outcomes. We therefore adopt a dual-mode prediction mechanism for inquiry and intervention that conditions the model behavior on the semantic type of each action.

\paragraph{Inquiry Mode.} \mbox{}
For inquiry actions $a_t^{(j)} \in \mathcal{A}_{\text{inq}}$, such as laboratory tests or physical examinations, the model predicts explicit observable values. The training objective for these actions is the negative log-likelihood of the ground-truth observations:
\begin{equation}
    \mathcal{L}_{\text{inq}} = - \sum_{j: a_t^{(j)} \in \mathcal{A}_{\text{inq}}}
    \log P_\theta(v_{\text{gt}}^{(j)} \mid S_t, \mathbf{A}_t).
\end{equation}

% \paragraph{Intervention Mode.} \mbox{}
% For intervention actions $a_t^{(j)} \in \mathcal{A}_{\text{int}}$, such as medication administrations or procedures, no immediate observable value is produced. We model this by assigning an empty outcome, indicating the absence of a direct observation:
% \begin{equation}
%     P_\theta(v_t^{(j)} = \emptyset) = 1.
% \end{equation}
% Although the intervention actions do not explicitly incur any direct supervision loss, they nevertheless function as indispensable control signals that  modulate the underlying patient state and influence the trajectory of all future clinical predictions.

\paragraph{Intervention Mode.} \mbox{}
For intervention actions $a_t^{(j)} \in \mathcal{A}_{\text{int}}$ (e.g., medications or procedures), the model is not required to generate an immediate observable value. We formalize this by assigning an empty outcome $P_\theta(v_t^{(j)} = \emptyset) = 1$. Instead of yielding direct feedback, these actions are recorded into the patient's history as state-altering intervention events. Their primary function is to update the simulation context, thereby influencing the trajectory of future clinical predictions.

\subsection{Deterministic State Transition}

After generating the outcome set $\mathbf{V}_t$, we construct the event set at step $t$ as $\mathbf{E}_t = \{(a_t^{(j)}, v_t^{(j)})\}_{j=1}^{N}$. 
Consequently, the interaction history is then deterministically updated:
\begin{equation}
\begin{aligned}
    \mathcal{H}_{t+1} &= \mathcal{H}_t \cup \{\mathbf{E}_t\}, \\
    S_{t+1} &= \langle \tau_{t+1}, d, \mathcal{Y}, \mathcal{H}_{t+1} \rangle.
\end{aligned}
\end{equation}
This update loop ensures that all subsequent predictions are conditioned on the complete sequence of prior inquiries and interventions, enabling consistent long-horizon simulation of disease progression under clinical decision-making.
Appendix~\ref{app:algorithm} provides a description of the simulation procedure.

\section{Experiments}
\label{sec:experiments}

\definecolor{groupbg}{HTML}{EAEAEA} % Define the gray color for group headers

\begin{table*}[t]
\centering
\small
\renewcommand{\arraystretch}{1.2}
\setlength{\tabcolsep}{4pt}

\caption{Overall performance of world model under full trajectory prediction. We evaluate numerical precision and label accuracy. Avg Score is calculated as the mean of S@25, Stat F1, and Label F1. Err denotes SMAPE. $\uparrow$ indicates higher is better, $\downarrow$ indicates lower is better. The best results are \textbf{bolded} and the second best are \underline{underlined}.}
\label{tab:full_trajectory_main}

\resizebox{0.98\textwidth}{!}
{
\begin{tabular}{l c ccccc ccc c}
\toprule
\multirow{2}{*}{\textbf{Model}} & \multirow{2}{*}{\textbf{Size}} & \multicolumn{5}{c}{\textbf{Numerical Precision}} & \multicolumn{3}{c}{\textbf{Label Accuracy}} & \multirow{2}{*}{\textbf{Avg}} \\
\cmidrule(lr){3-7} \cmidrule(lr){8-10} 
 & & S@10 $\uparrow$ & S@15 $\uparrow$ & S@25 $\uparrow$ & Err $\downarrow$ & Stat F1 $\uparrow$ & Precision $\uparrow$ & Recall $\uparrow$ & F1 $\uparrow$ & Score $\uparrow$ \\
\midrule

%----------------------------------------------------------------
% Group 1: Closed-Source Models
%----------------------------------------------------------------
\rowcolor{groupbg}
\multicolumn{11}{c}{\textit{\textbf{Closed-Source Models}}} \\
\textit{GPT-5.2}                     & $-$ & $0.435$ & $0.540$ & $0.674$ & $0.348$ & $0.627$ & $0.635$ & $0.490$ & $0.553$ & $0.618$ \\
\textit{Gemini-3.0-Pro-Preview}      & $-$ & $0.445$ & $0.543$ & $0.681$ & $0.299$ & $0.565$ & $0.481$ & $0.473$ & $0.477$ & $0.574$ \\

\midrule
%----------------------------------------------------------------
% Group 2: Open-Source General Models
%----------------------------------------------------------------
\rowcolor{groupbg}
\multicolumn{11}{c}{\textit{\textbf{Open-Source General-Purpose Models}}} \\
\textit{Qwen3-4B-Instruct}           & 4B   & $0.365$ & $0.471$ & $0.607$ & $0.448$ & $0.473$ & $0.450$ & $0.258$ & $0.328$ & $0.469$ \\
\textit{Qwen3-8B}                    & 8B   & $0.359$ & $0.463$ & $0.600$ & $0.437$ & $0.466$ & $0.448$ & $0.373$ & $0.407$ & $0.491$ \\
\textit{Qwen2.5-14B-Instruct}        & 14B  & $0.393$ & $0.507$ & $0.659$ & $0.339$ & $0.423$ & $0.358$ & $0.237$ & $0.285$ & $0.456$ \\
\textit{Qwen3-30B-A3B-Instruct}      & 30B  & $0.403$ & $0.508$ & $0.645$ & $0.338$ &   $0.419$ & $0.324$ & $0.415$ & $0.364$ & $0.476$ \\
\textit{Llama-3.3-70B-Instruct}      & 70B  & $0.381$ & $0.484$ & $0.625$ & $0.364$ & $0.377$ & $0.581$ & $0.439$ & $0.500$ & $0.501$ \\
\textit{Qwen3-Next-80B-A3B-Instruct} & 80B  & $0.398$ & $0.501$ & $0.630$ & $0.381$ & $0.538$ & $0.416$ & $0.471$ & $0.442$ & $0.537$ \\
\textit{GPT-OSS-120B}                & 120B & $0.402$ & $0.502$ & $0.634$ & $0.372$ & $0.483$ & $0.432$ & $0.346$ & $0.384$ & $0.500$ \\
\textit{MiniMax-M2.1}                & 229B & $0.410$ & $0.513$ & $0.650$ & $0.338$ & $0.502$ & $0.475$ & $0.486$ & $0.480$ & $0.544$ \\
\textit{Qwen3-235B-A22B-Instruct}    & 235B & $0.409$ & $0.514$ & $0.647$ & $0.339$ & $0.469$ & $0.483$ & $0.343$ & $0.402$ & $0.506$ \\
\textit{GLM-4.7}                     & 358B & $0.442$ & $0.525$ & $0.663$ & $0.322$ & $0.476$ & $0.381$ & $0.254$ & $0.305$ & $0.481$ \\
\textit{DeepSeek-V3.2}               & 671B & $0.416$ & $0.518$ & $0.649$ & $0.340$ & $0.554$ & $0.509$ & $0.402$ & $0.449$ & $0.551$ \\

\midrule
%----------------------------------------------------------------
% Group 3: Open-Source Medical Models
%----------------------------------------------------------------
\rowcolor{groupbg}
\multicolumn{11}{c}{\textit{\textbf{Open-Source Medical Models}}} \\
\textit{MedGemma-4B-IT}    & 4B  & $0.336$ & $0.430$ & $0.554$ & $0.561$ & $0.438$ & $0.412$ & $0.142$ & $0.211$ & $0.401$ \\
\textit{MedGemma-27B-IT}   & 27B & $0.364$ & $0.460$ & $0.588$ & $0.496$ & $0.436$ & $0.259$ & $0.205$ & $0.229$ & $0.418$ \\
\textit{Baichuan-M2-32B}   & 32B & $0.372$ & $0.472$ & $0.601$ & $0.418$ & $0.499$ & $0.456$ & $0.386$ & $0.418$ & $0.506$ \\

\midrule
%----------------------------------------------------------------
% Group 4: Ours
%----------------------------------------------------------------
\rowcolor{groupbg}
\multicolumn{11}{c}{\textit{\textbf{Ours ( EHRWorld)}}} \\
\rowcolor{oursbg}
\textit{\textbf{ EHRWorld-4B}}  & 4B  & $0.460$ & $0.566$ & $0.703$ & $0.274$ & $0.649$ & $\mathbf{0.939}$ & $0.886$ & $\underline{0.912}$ & $0.755$ \\
\rowcolor{oursbg}
\textit{\textbf{ EHRWorld-8B}}  & 8B  & $\underline{0.468}$ & $\underline{0.577}$ & $\underline{0.714}$ & $\underline{0.269}$ & $\underline{0.658}$ & $\underline{0.936}$ & $\underline{0.891}$ & $\mathbf{0.913}$ & $\underline{0.762}$ \\
\rowcolor{oursbg}
\textit{\textbf{ EHRWorld-14B}} & 14B & $\mathbf{0.475}$ & $\mathbf{0.582}$ & $\mathbf{0.716}$ & $\mathbf{0.262}$ & $\mathbf{0.667}$ & $0.925$ & $\mathbf{0.901}$ & $\mathbf{0.913}$ & $\mathbf{0.765}$ \\
\bottomrule
\end{tabular}
}
\end{table*}

\subsection{Experimental Setup}
\label{sec:setup}

\noindent\textbf{Implementation Details.}
We train three medical world model variants, \emph{ EHRWorld-4B}, \emph{ EHRWorld-8B}, and \emph{ EHRWorld-14B}, by fine-tuning the \emph{Qwen3-4B-Instruct}, \emph{Qwen3-8B}~\cite{qwen3}, and \emph{Qwen2.5-14B-Instruct}~\cite{qwen2.5} foundation models on the  EHRWorld-110K dataset. All models are trained for three epochs.
Training is performed on a cluster of $8\times$ NVIDIA H100 GPUs using DeepSpeed ZeRO-2 optimization~\cite{rasley2020deepspeed}. We adopt AdamW~\cite{loshchilov2019decoupledweightdecayregularization} with a learning rate of $1\times10^{-6}$, a cosine decay schedule, and a warm-up ratio of 0.05. During training, models are optimized to predict outcomes conditioned on sets of clinical events, with causal masking applied to preserve the temporal order.

\noindent\textbf{Baselines.}
We compare our models against a diverse set of strong LLMs baselines to assess performance across different model scales and training paradigms. The baselines are grouped into three categories based on availability and domain specialization:
\textbf{(1) Closed-Source Models}, including the proprietary frontier systems \emph{GPT-5.2}~\cite{openai2025gpt5systemcard} and \emph{Gemini-3.0-Pro-Preview}~\cite{gemini2023};
\textbf{(2) Open-Source General-Purpose Models}, spanning a wide range of parameter scales, including \emph{Qwen3-4B-Instruct}, \emph{Qwen3-8B}, \emph{Qwen2.5-14B-Instruct}, and \emph{Qwen3-30B-A3B-Instruct} as well as larger models such as \emph{Llama-3.3-70B-Instruct}~\cite{dubey2024llama}, \emph{Qwen3-Next-80B-A3B-Instruct}, \emph{GPT-OSS-120B}~\cite{openai2025gptoss120bgptoss20bmodel}, \emph{MiniMax-M2.1}~\cite{minimax2025minimaxm1scalingtesttimecompute}, \emph{Qwen3-235B-A22B-Instruct}, \emph{GLM-4.7}~\cite{5team2025glm45agenticreasoningcoding}, and \emph{DeepSeek-V3.2}~\cite{deepseekai2025deepseekv32};
and \textbf{(3) Open-Source Medical Models}, which are typically trained or adapted on static medical corpora such as textbooks, clinical guidelines, and case reports. This category includes \emph{MedGemma-4B-IT}, \emph{MedGemma-27B-IT}~\cite{sellergren2025medgemma}, and \emph{Baichuan-M2-32B}~\cite{dou2025baichuan}.
All baseline models are evaluated under their default inference configurations without task-specific fine-tuning.

\noindent\textbf{Evaluation Metrics.}
We evaluate the fidelity of simulated physiological states using a multi-dimensional metric suite, with formal definitions of all metrics provided in Appendix~\ref{app:metrics}. Our evaluation considers two complementary aspects: 
(1) \textbf{Numerical Precision} for continuous variables (e.g., vital signs and laboratory results). Numerical precision is assessed using regression-based metrics, including Success Rate (S@k) and SMAPE. Beyond raw numerical error, we further evaluate clinical interpretability by mapping predicted values to normal or abnormal ranges defined by clinical reference standards, and report Clinical Status F1 to measure whether pathological deviations are correctly captured. 
(2) \textbf{Label Accuracy} for discrete outputs (e.g., microbiology results), where label correctness is evaluated using standard classification metrics, including precision, recall, and F1 score.

\begin{table*}[t]
\centering
\small
\renewcommand{\arraystretch}{1.25} 
\setlength{\tabcolsep}{2.5pt}      

\caption{Stability Analysis: \textbf{Next-Step Prediction} vs. \textbf{Full Trajectory Prediction}. We assess model robustness across three dimensions: Numerical Precision (Numer. S@25), Clinical Interpretation (Stat F1), and Label Accuracy (Label F1). Ret (\%) denotes the Retention Rate ($\frac{\text{Full}}{\text{Next}} \times 100$), quantifying the model's resistance to error accumulation during long-horizon generation. The best results are \textbf{bolded} and the second best are \underline{underlined}.} 
\label{tab:stability_analysis}

\resizebox{0.98\textwidth}{!}{
\begin{tabular}{l ccc ccc ccc ccc}
\toprule
% ================= Header Level 1 =================
\multirow{2}{*}{\textbf{Model}} & 
\multicolumn{3}{c}{\textbf{Numer. S@25}} & 
\multicolumn{3}{c}{\textbf{Stat F1}} & 
\multicolumn{3}{c}{\textbf{Label F1}} & 
\multicolumn{3}{c}{\textbf{Overall}} \\

% Header Rules (Grouping)
\cmidrule(lr){2-4} \cmidrule(lr){5-7} \cmidrule(lr){8-10} \cmidrule(lr){11-13}

% ================= Header Level 2 =================
 & Next & Full & Ret (\%) $\uparrow$ & Next & Full & Ret (\%) $\uparrow$ & Next & Full & Ret (\%) $\uparrow$ & Next & Full & Ret (\%) $\uparrow$ \\
\midrule
\textit{Baichuan-M2-32B}             & $0.752$ & $0.601$ & $79.9$ & $0.692$ & $0.499$ & $72.1$ & $0.527$ & $0.418$ & $79.3$ & $0.657$ & $0.506$ & $73.4$ \\
\textit{Qwen3-Next-80B-A3B-Instruct} & $0.759$ & $0.630$ & $83.0$ & $0.715$ & $0.538$ & $75.2$ & $0.558$ & $0.442$ & $79.2$ & $0.677$ & $0.537$ & $76.3$ \\
\textit{Qwen3-235B-A22B-Instruct}    & $0.769$ & $0.647$ & $84.1$ & $0.707$ & $0.469$ & $66.3$ & $0.557$ & $0.402$ & $72.2$ & $0.678$ & $0.506$ & $71.9$ \\
\textit{GLM-4.7}                     & $0.775$ & $0.663$ & $\underline{85.5}$ & $0.670$ & $0.476$ & $71.0$ & $0.426$ & $0.305$ & $71.6$ & $0.624$ & $0.481$ & $77.1$ \\
\textit{DeepSeek-V3.2}               & $0.770$ & $0.649$ & $82.3$ & $0.722$ & $0.554$ & $76.7$ & $0.608$ & $0.449$ & $73.8$ & $0.700$ & $0.551$ & $77.6$ \\
\textit{GPT-5.2}                     & $\underline{0.789}$ & $\underline{0.674}$ & $85.4$ & $\underline{0.741}$ & $\underline{0.627}$ & $\underline{84.6}$ & $\underline{0.620}$ & $\underline{0.553}$ & $\underline{86.2}$ & $\underline{0.717}$ & $\underline{0.618}$ & $\underline{86.2}$ \\
\rowcolor{oursbg}
\textit{\textbf{ EHRWorld-14B (Ours)}} & $\mathbf{0.806}$ & $\mathbf{0.716}$ & $\mathbf{88.8}$ & $\mathbf{0.784}$ & $\mathbf{0.667}$ & $\mathbf{85.1}$ & $\mathbf{0.928}$ & $\mathbf{0.913}$ & $\mathbf{98.4}$ & $\mathbf{0.839}$ & $\mathbf{0.765}$ & $\mathbf{92.6}$ \\

\bottomrule
\end{tabular}
}
\end{table*}
% \vspace{0.5em}

\subsection{Main Results: Full Trajectory Prediction}
\label{sec:exp_full_trajectory}

A robust medical world model must demonstrate the ability to simulate the continuous evolution of patient health over extended horizons, maintaining coherence under sequential interventions. To test this capability, we evaluate full trajectory prediction, where the model must autoregressively generate the entire hospitalization trajectory from admission to discharge, conditioned on sequential clinical events and interventions. Table \ref{tab:full_trajectory_main} presents a comprehensive comparison between our  EHRWorld family and a diverse set of baselines.

% As aforementioned in Section~\ref{sec:intro}, naive LLM-based simulators exhibit pronounced error drift in long-horizon clinical trajectory, leading to degraded downstream state consistency. This phenomenon persists even for strong closed-source baselines: \textit{GPT-5.2} and \textit{Gemini-3.0-Pro-Preview} achieve Avg Scores of $0.618$ and $0.574$, respectively. In contrast,  EHRWorld attains an Avg Score of $0.755$--$0.765$, yielding an absolute improvement of up to $+0.147$ over the best closed-source model. These results demonstrate that \textbf{ EHRWorld yields substantially more stable long-horizon simulation with reduced error drift}, maintaining more coherent state evolution under sequential interventions.

As aforementioned in Section~\ref{sec:intro}, naive LLM-based simulators exhibit pronounced error drift in long-horizon clinical trajectory, leading to degraded downstream state consistency. This phenomenon persists even for strong closed-source baselines: \textit{GPT-5.2} and \textit{Gemini-3.0-Pro-Preview} achieve Avg Scores of $0.618$ and $0.574$, respectively. In contrast,  EHRWorld attains an Avg Score of $0.755$--$0.765$. These results demonstrate that \textbf{EHRWorld yields substantially more stable long-horizon simulation with reduced error drift}.

% The performance gap observed in Table~\ref{tab:full_trajectory_main} further indicates that parameter scaling alone does not reliably improve long-horizon trajectory simulation for general-purpose LLMs. For instance, despite its massive scale, \textit{DeepSeek-V3.2} attains an Avg Score of only $0.551$, which remains significantly lower than our models. Notably, even \textit{EHRWorld-4B} achieves stronger numerical precision (S@25=$0.703$) with lower error (Err=$0.274$) than much larger baselines such as \textit{Gemini-3.0-Pro-Preview} (S@25=$0.681$, Err=$0.299$). This disparity underscores that \textbf{causal sequential training matters more than parameter scaling for modeling patient dynamics}, as specialized training objectives better capture the underlying temporal dependencies than simple scale.

\begin{figure}[t]
    \centering
    \includegraphics[width=\linewidth]{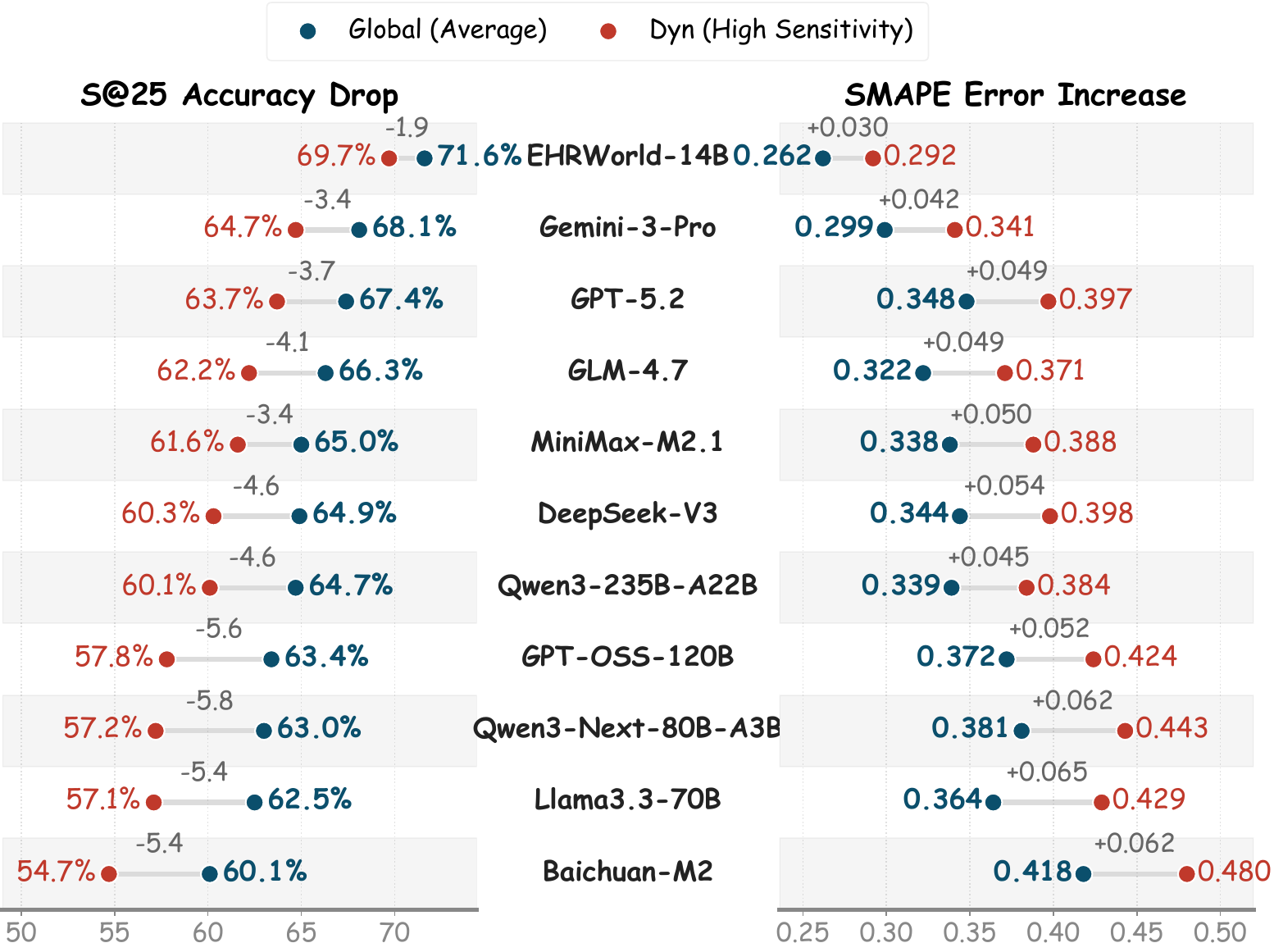}
    \caption{Performance stability analysis across global and dynamic clinical metrics. \textit{EHRWorld-14B} demonstrates superior robustness, exhibiting the smallest performance degradation in S@25 accuracy and SMAPE error compared to general foundation models.}
\label{fig:dynamic_stability}
\end{figure}
\vspace{0.5em}

The performance gap further indicates that parameter scaling alone does not reliably improve long-horizon trajectory simulation for general-purpose LLMs. For instance, scaling from \textit{Qwen3-30B-A3B-Instruct} to the massive \textit{Qwen3-235B-A22B-Instruct} yields negligible gains in numerical precision, with S@25 increasing marginally from $0.645$ to $0.647$. In stark contrast, even our smallest model, \textit{EHRWorld-4B}, achieves a superior S@25 of $0.703$, outperforming much larger baselines including the closed-source \textit{Gemini-3.0-Pro-Preview} ($0.681$). This disparity underscores that \textbf{causal sequential training matters more than parameter scaling for modeling patient dynamics}, as specialized training objectives better capture underlying temporal dependencies than simple scale.

% Furthermore, we observe that \textbf{medical specialization on static corpora is insufficient for long-horizon trajectory prediction.} While models specialized via medical instruction tuning excel at tasks like medical question-answering and report generation, they still struggle with sustained clinical rollouts. As shown in Table~\ref{tab:full_trajectory_main}, medical baselines such as \textit{MedGemma} and \textit{Baichuan-M2} achieve lower Avg Scores ($0.401$--$0.506$), lagging behind general-purpose models and falling significantly below EHRWorld ($0.755$--$0.765$). These results suggest that training on static knowledge does not naturally translate into reliable sequential simulation, reinforcing the need for longitudinal, intervention-conditioned supervision for dynamic modeling.

Furthermore, we observe that \textbf{medical LLM is insufficient for long-horizon trajectory prediction.} While models specialized via medical instruction tuning excel at tasks like medical question-answering and report generation, they still struggle with sustained clinical rollouts. As shown in Table~\ref{tab:full_trajectory_main}, medical baselines such as \textit{MedGemma} and \textit{Baichuan-M2} achieve lower Avg Scores ($0.401$--$0.506$), lagging behind general-purpose models and falling significantly below EHRWorld ($0.755$--$0.765$). These results suggest that training on static knowledge does not naturally translate into reliable sequential simulation, reinforcing the need for longitudinal, intervention-conditioned supervision for dynamic modeling.

\subsection{Next-Step vs. Full Trajectory Prediction}
\label{sec:exp_stability}

\begin{figure*}[t]
    \centering
    \includegraphics[width=0.98\textwidth]{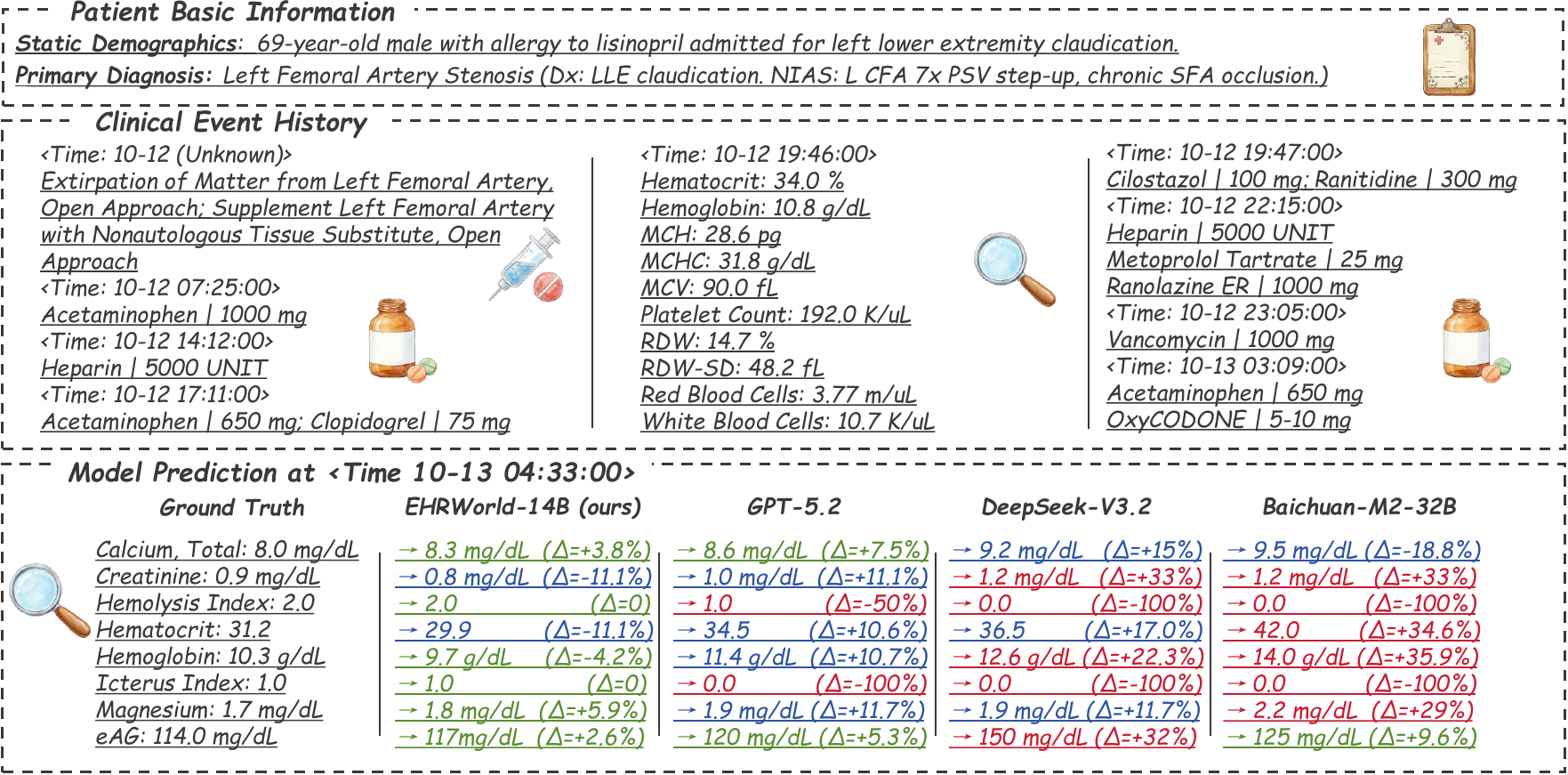}
    \caption{
\textbf{Qualitative case study comparison.} 
We visualize the ground truth versus model predictions for a 69-year-old patient. Colors denote relative error ($\Delta$) severity: \textcolor{c_good}{\textbf{precise}} ($\Delta \le 10\%$), \textcolor{c_mid}{\textbf{acceptable}} ($10\% < \Delta \le 20\%$), and \textcolor{red}{\textbf{deviation}} ($\Delta > 20\%$). \textit{EHRWorld-14B} shows superior stability, keeping most metrics within low-error margin.
}
\label{fig:case_study}
\end{figure*}

We evaluate model performance in two distinct settings: next-step prediction, where the model predicts the state at $t\!+\!1$ conditioned on ground-truth history, and full-trajectory simulation, where predictions are generated autoregressively. This isolates the model's intrinsic predictive capability from the effects of compounding rollout errors. To quantify robustness against error accumulation, we introduce the Retention Rate, defined as $\text{Ret}(\%)=\frac{\text{Full}}{\text{Next}}\times 100$, where a higher percentage indicates greater resilience and stability over long horizons.

The results in Table~\ref{tab:stability_analysis} reveal a critical distinction between local forecasting and long-term stability. While the performance gap between baselines and EHRWorld is narrow in next-step prediction (e.g., Numer. S@25 ranges tightly from $0.752$ to $0.806$), it widens significantly in the full-trajectory setting. \textit{EHRWorld-14B} achieves a superior overall retention rate of $92.6\%$, surpassing the best baseline, \textit{GPT-5.2} ($86.2\%$). This indicates that while many general-purpose LLMs can effectively predict immediate outcomes when teacher-forced, they fail to maintain coherent patient states over time. In contrast, EHRWorld demonstrates exceptional stability, effectively mitigating the error accumulation that degrades performance in standard models.

\subsection{High-Sensitivity Stability Analysis}
\label{sec:dyn_stability}

Clinical trajectories often contain abrupt physiological shifts (e.g., acute deterioration or rapid treatment response), where small inconsistencies can be amplified into clinically implausible rollouts. To stress-test robustness under such transitions, we construct a \emph{high-sensitivity} subset in which the relative change between two consecutive steps exceeds $50\%$. We then compare model performance on the global test distribution against this high-sensitivity subset, reporting the degradation in S@25 (accuracy drop) and SMAPE (error increase). 

As shown in Figure~\ref{fig:dynamic_stability}, \textit{EHRWorld-14B} shows the most stable behavior under high-sensitivity shifts, with S@25 decreasing from $71.6\%$ to $69.7\%$ (a $-1.9$ point drop) and SMAPE increasing only by $+0.030$ ($0.262\!\rightarrow\!0.292$). In comparison, general foundation models exhibit larger accuracy drops and more pronounced SMAPE inflation on the same subset, suggesting that their predictions are more susceptible to drift when the patient state changes abruptly. Overall, \textit{EHRWorld-14B}'s smaller gap between global and high-sensitivity performance indicates stronger consistency in tracking rapidly evolving clinical states.

% \begin{figure}[t]
%     \centering
%     \includegraphics[width=\linewidth]{images/time_compare.pdf}
%     \caption{Comparison of inference time and token throughput across different models. \textit{EHRWorld-14B} demonstrates superior computational efficiency, achieving the fastest generation speed while maintaining the highest volume compared to all baselines.}
%     \label{fig:efficiency}
% \end{figure}

% \vspace{0.5em}

% \subsection{Inference Efficiency Analysis}
% \label{sec:efficiency}

% To assess deployment practicality for large-scale clinical simulation, we benchmark inference efficiency on a cluster with $8\times$ NVIDIA H100 GPUs. We restrict our comparison to competitive models that are viable for deployment on this specific hardware configuration. All models are served with SGLang~\cite{zheng2024sglangefficientexecutionstructured} under a request concurrency of 64, and we report average latency per case together with aggregate token throughput.As shown in Figure~\ref{fig:efficiency}, \textit{EHRWorld-14B} achieves the best efficiency profile, exhibiting the lowest latency while delivering the highest throughput among all evaluated candidates. It remains consistently faster than both substantially larger general-purpose models and domain-specific medical baselines.

% Additional case studies are  in Appendix~\ref{app:case}.

\subsection{Case Study}

Figure~\ref{fig:case_study} shows the trajectory prediction for a patient with left femoral artery stenosis. We categorize prediction accuracy into three levels: precise ($\le$10\%), acceptable (10--20\%), and deviation ($>$20\%). \textit{EHRWorld-14B} performs well, with most predictions falling in the precise range. For more dynamic metrics, such as \textit{Hematocrit}, it maintains accuracy within the acceptable range, outperforming other models. This demonstrates EHRWorld's ability to maintain state consistency over time, unlike models that tend to deviate.

\section{Conclusion}
\label{sec:conclusion}

In this study, we evaluated the ability of LLMs to function as medical world models for long-horizon clinical simulations. We found that models incorporating medical knowledge struggle with state inconsistency and error drift under sequential interventions. To address this, we introduced EHRWorld and the EHRWorld-110K dataset, enabling causally-aligned modeling of clinical trajectories. Extensive experiments show that EHRWorld outperforms both proprietary and open-source baselines in long-horizon clinical trajectory prediction.

%We evaluated the ability of LLMs as medical world models for long-horizon clinical simulations. Our findings show that models trained on static medical corpora struggle with state inconsistency and error drift. We introduced EHRWorld and the EHRWorld-110K dataset, which enables better modeling of hospitalization trajectories. Experiments demonstrate that EHRWorld outperforms baselines in full-trajectory prediction, highlighting the importance of intervention-conditioned supervision for coherent clinical dynamics.

\section*{Limitations}
Despite its strong performance, EHRWorld has several limitations. First, although the proposed training paradigm substantially reduces long-horizon error accumulation, the model still relies on autoregressive generation, which can propagate residual inaccuracies over extended rollouts. Second, our evaluation primarily focuses on trajectory-level fidelity, stability, and consistency under sequential interventions, and does not directly assess downstream clinical utility such as treatment optimization, policy learning, or real-world decision support. Future work could extend the evaluation to decision-centric settings, including counterfactual treatment comparison and clinician-in-the-loop studies, to better understand the practical implications of medical world models.

\section*{Ethical Considerations}
This work involves modeling sensitive clinical processes and therefore raises important ethical considerations. All data used to construct EHRWorld-110K are de-identified and processed in compliance with established privacy and data protection standards. However, models trained on observational clinical data may reflect existing practice patterns and implicit biases, which could influence simulated outcomes if interpreted without caution. We emphasize that EHRWorld is designed as a research-oriented medical world model for simulation and analysis, rather than a system intended for autonomous or real-time clinical decision-making. Any application beyond research settings would require careful validation, transparency, and human oversight.
\bibliography{custom}

\clearpage
\newpage
% \onecolumn
\appendix

\section{Details on Data Construction}
\label{app:data_details}

In this section, we provide a comprehensive breakdown of the data construction process for the EHRWorld-110K dataset, focusing on the extraction of static patient context and the statistical characteristics of the processed clinical trajectories.

\subsection{Static Information Extraction}

To obtain a structured representation of patient demographics and clinical context from unstructured textual records, we employed the \emph{Qwen3-235B-A22B-Instruct}~\cite{qwen3}. The extraction process targeted raw discharge summaries from MIMIC-IV to identify key static attributes, including age, gender, allergy history, and a hierarchical set of diagnoses (primary vs. secondary).

We designed a specific system prompt to guide the LLM in parsing these narratives into a standardized JSON schema. The prompt strictly constrains the model to extract only diagnoses relevant to the current hospital admission, filtering out historical conditions unless they were actively treated. The full prompt template used in our pipeline is presented below:

% --- Prompt Block Starts Here ---
\begin{tcolorbox}[
    enhanced,
    breakable, 
    title=\textbf{Static Information Extraction},
    colframe=promptframe, % 确保你在导言区定义了颜色，或者改成 black!70
    colback=promptbg,     % 确保你在导言区定义了颜色，或者改成 gray!5
    coltitle=white,      
    fonttitle=\bfseries,
    attach boxed title to top left={xshift=5mm, yshift=-2mm},
    boxed title style={size=small, colback=promptframe},
    boxrule=0.5mm,
    top=4mm,
    % drop shadow medium % [已注释] 防止报错，如需阴影请在导言区加载 \tcbuselibrary{skins}
]
    \small 
    \textbf{\# Role}\\
    You are an expert Clinical Data Analyst and Medical Terminology Specialist. Your task is to extract specific data points from a patient's \textbf{Discharge Record} and structure them into a precise JSON format.

    \vspace{0.5em}
    \textbf{\# Instructions}\\
    Analyze the provided text and extract the information into the fields defined below.

    \begin{itemize}[leftmargin=1.5em, noitemsep, topsep=2pt]
        \item \textbf{1. Basic Information}: Extract Age, Gender, Allergy History, and Chief Complaint.
        \item \textbf{2. History Information}: Summarize PMH, Family History, and Social History.
        \item \textbf{3. Diagnosis Results}: 
        \begin{itemize}[leftmargin=1em]
            \item Extract ONLY diagnoses addressed during \textbf{this stay}.
            \item Split into \textit{Primary Diagnosis} and \textit{Secondary Diagnoses} (sorted chronologically).
        \end{itemize}
    \end{itemize}

    \vspace{0.5em}
    \textbf{\# Output Format (JSON Template)}
    \lstset{
        basicstyle=\ttfamily\scriptsize,
        breaklines=true,
        columns=fullflexible,
        moredelim=[s][\color{jsonkey}]{"}{:},
        stringstyle=\color{jsonstring}
    }
    \begin{lstlisting}
{
  "Basic Information": "String...",
  "History Information": "String...",
  "Diagnosis Results": {
    "Primary Diagnosis": { "Content": "...", "Reason": "..." },
    "Secondary Diagnoses": [
      { "Content": "...", "Reason": "..." }
    ]
  }
}
    \end{lstlisting}
\end{tcolorbox}
% --- Prompt Block Ends Here ---

\subsection{Data Filtering and Partitioning}

To ensure the robustness of the benchmark and mitigate the noise inherent in raw electronic health records, we implemented a rigorous two-stage data processing pipeline consisting of statistical filtering and patient-centric partitioning.

\paragraph{Statistical Filtering and Outlier Removal.} 
Clinical event sequences often exhibit long-tail distributions that introduce computational inefficiency and modeling instability. To address this, we first computed the population-level statistics for key metrics, including Length of Stay (LOS) and the frequency of distinct event types (laboratory tests, microbiology, and medication administrations). We applied a truncation strategy where any admission episode containing a metric exceeding the 90th percentile was excluded. Furthermore, we restricted the total event count per episode to the 10th--90th percentile range, removing sequences that were either uninformatively short or excessively long. To guarantee that all samples contained meaningful therapeutic signals, we strictly filtered out episodes with zero medication administration (EMAR) records.

\paragraph{Patient-Centric Splitting Strategy.} 
To rigorously evaluate the model's ability to generalize to unseen subjects, we adopted a \textit{patient-centric} partitioning strategy rather than a random admission-level split. We grouped all data by unique patient identifiers (\texttt{subject\_id}) and assigned patients exclusively to either the training or the test set. This ensures that all hospitalization episodes for a given individual reside in the same partition, thereby preventing data leakage where the model might otherwise exploit patient-specific idiosyncrasies or historical correlations across different admissions.

\begin{figure*}[t]
    \centering
    \includegraphics[width=\linewidth]{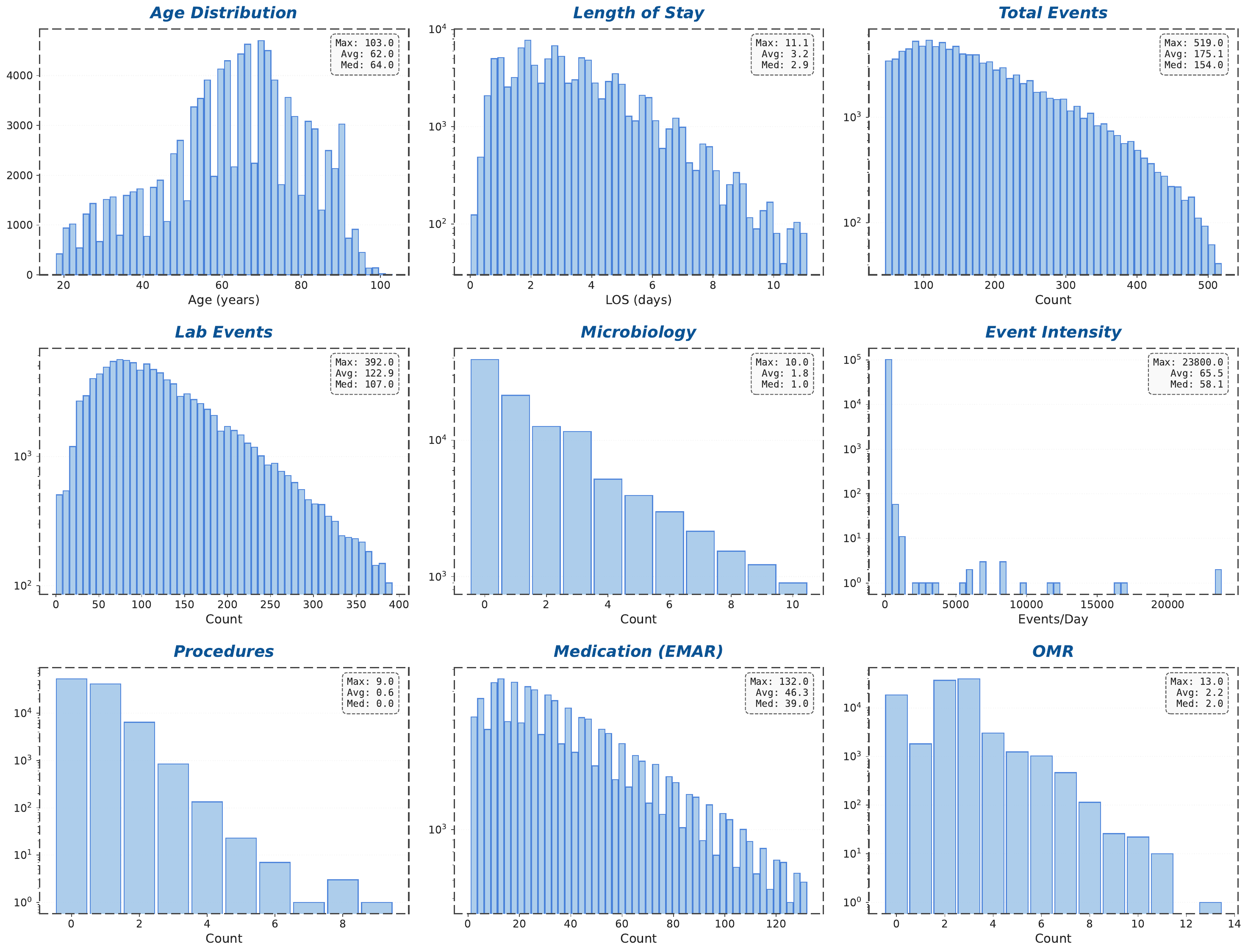}
    \caption{Distributions of demographics and clinical event statistics for the processed EHRWorld-110K dataset. The histograms illustrate patient age, length of stay (LOS), and event counts across diverse modalities (e.g., Lab Events, Medications, Procedures). Additionally, we plot Event Intensity (total events divided by LOS) to represent the density of clinical activities. Inset boxes report descriptive statistics (maximum, mean, and median) for each metric. Note that the y-axes for LOS and event counts use a logarithmic scale to visualize the long-tail distributions.}
    \label{fig:efficiency}
\end{figure*}

\subsection{Data Statistics and Distribution}
\label{sec:data_stats}

The \textit{EHRWorld-110K} dataset exhibits diverse distributions across key demographic and clinical metrics, as shown in Figure~\ref{fig:efficiency}. These distributions are critical for characterizing the variability and clinical fidelity of the data.

\paragraph{Patient Age.} 
The dataset spans a broad age spectrum, with a predominant concentration of patients between 40 and 80 years old. A pronounced peak is observed in the 50--60 age bracket, reflecting a significant representation of middle-aged and elderly patients, which is typical for hospital-based cohorts.

\paragraph{Length of Stay (LOS).} 
The LOS distribution follows a significant long-tail pattern. While the majority of hospitalizations are relatively brief (1--3 days), a subset of patients requires extended care, with some episodes exceeding 100 days. This skewed distribution is visualized using a logarithmic scale to effectively capture the contrast between routine short-term admissions and protracted hospitalization episodes.

\paragraph{Clinical Event Counts.} 
Event frequencies vary significantly by modality; laboratory tests and medication administrations are the most voluminous, whereas procedural events are comparatively sparse. Similar to LOS, these counts exhibit a heavy-tailed distribution: while most episodes involve moderate clinical activity, a distinct fraction of patients undergoes extensive interventions, accumulating high volumes of diagnostic and therapeutic records.

\paragraph{Event Intensity.} 
Event Intensity, defined as the total number of clinical events normalized by the LOS, serves as a proxy for the density of medical care. This metric reveals that episodes characterized by high event intensity are often associated with longer hospitalizations, indicating that patients requiring prolonged treatment also demand more frequent monitoring and intervention.

The descriptive statistics (mean, median, and maximum) presented in the inset boxes of Figure~\ref{fig:efficiency} further quantify this variability. For instance, the divergence between the mean LOS ($\approx$ 5 days) and the maximum LOS ($>$ 100 days) underscores the wide range of patient care trajectories. Collectively, these statistics highlight the heterogeneity of the \textit{EHRWorld-110K} dataset, providing a rich and varied foundation for evaluating models in clinical data analysis.

\section{EHRWorld Simulation Algorithm}
\label{app:algorithm}

This appendix provides a procedural description of the  simulation process to complement the formulation in Section~\ref{sec:model}. The algorithm summarizes how patient states are updated over discrete simulation steps given sets of clinical actions, following the dual-mode prediction mechanism for inquiry and intervention events. The procedure is intended to clarify the execution flow of the world model and does not introduce additional modeling assumptions beyond those described in the main text.

\begin{algorithm}
\caption{Sequential Simulation}
\label{alg:EHRWorld}
\KwIn{
Initial patient context $(d, \mathcal{Y})$, initial timestamp $\tau_1$, trained model $P_\theta$
}
\KwOut{
Simulated clinical trajectory $\mathcal{H}$
}

Initialize $\mathcal{H}_1 \leftarrow \emptyset$\;
Initialize state $S_1 \leftarrow \langle \tau_1, d, \mathcal{Y}, \mathcal{H}_1 \rangle$\;

\For{$t = 1, 2, \dots$}{
    Receive a set of clinical actions $\mathbf{A}_t = \{a_t^{(1)}, \dots, a_t^{(N)}\}$\;

    \ForEach{$a_t^{(j)} \in \mathbf{A}_t$}{
        \uIf{$a_t^{(j)} \in \mathcal{A}_{\text{inq}}$}{
            Sample observable outcome $v_t^{(j)} \sim P_\theta(v \mid S_t, \mathbf{A}_t)$\;
        }
        \ElseIf{$a_t^{(j)} \in \mathcal{A}_{\text{int}}$}{
            Set $v_t^{(j)} \leftarrow \emptyset$\;
        }
    }

    Construct event set $\mathbf{E}_t \leftarrow \{(a_t^{(j)}, v_t^{(j)})\}_{j=1}^{N}$\;

    Update history $\mathcal{H}_{t+1} \leftarrow \mathcal{H}_t \cup \{\mathbf{E}_t\}$\;

    Update state $S_{t+1} \leftarrow \langle \tau_{t+1}, d, \mathcal{Y}, \mathcal{H}_{t+1} \rangle$\;
}

\Return $\mathcal{H}$
\end{algorithm}

\section{Detailed Evaluation Metrics}
\label{app:metrics}

This appendix presents the formal mathematical definitions and methodologies employed to evaluate the fidelity of the \textbf{EHRWorld} World Model. Our evaluation framework is bifurcated into numerical precision for continuous physiological variables and semantic accuracy for discrete clinical events.

\subsection{Numerical Metrics}
To assess the generation of continuous values (e.g., vital signs, laboratory results), we utilize two complementary metrics that evaluate point-wise accuracy and robustness to numerical instability. Let $\mathcal{D} = \{(y_i, \hat{y}_i)\}_{i=1}^N$ denote a set of $N$ pairs consisting of ground truth values $y_i$ and model predictions $\hat{y}_i$.

\paragraph{1. Success Rate at Threshold $X$ (S@$X$).}
This metric quantifies the proportion of generated values falling within a clinically acceptable relative error margin $X$ (e.g., 10\%, 25\%). It serves as a direct proxy for clinical utility. We first compute the relative error $E_i$ for each sample:
\begin{equation}
    E_i = \left| \frac{y_i - \hat{y}_i}{y_i} \right|
\end{equation}
The Success Rate is subsequently defined as:
\begin{equation}
    \text{S@}X = \frac{1}{N} \sum_{i=1}^{N} \mathbb{I}\left(E_i \le \frac{X}{100}\right)
\end{equation}
where $\mathbb{I}(\cdot)$ represents the indicator function. We report \textbf{S@10} (Strict), \textbf{S@15} (Intermediate), and \textbf{S@25} (Standard). Higher values denote superior accuracy.

\paragraph{2. Symmetric Mean Absolute Percentage Error (SMAPE).}
Standard MAPE can be unstable when ground truth values approach zero and asymmetrically penalizes over-predictions. To mitigate this, we employ SMAPE (denoted as \textbf{Err} in experimental results), which ensures symmetric penalties and numerical stability:
\begin{equation}
    \text{SMAPE} = \frac{100\%}{N} \sum_{i=1}^{N} \frac{2 \cdot |y_i - \hat{y}_i|}{|y_i| + |\hat{y}_i| + \epsilon}
\end{equation}
where $\epsilon = 10^{-10}$ is a smoothing term included to prevent division by zero. Lower values indicate higher numerical precision.

\subsection{High-Sensitivity Analysis}
Clinical trajectories frequently involve rapid physiological shifts rather than steady states. Global aggregate metrics may obscure model performance during these critical episodes. To evaluate the model's responsiveness to acute changes, we identify a subset of high-sensitivity events, denoted as $\mathcal{H}_{sens}$.

An event at time $t$ is included in this subset if the ground truth value exhibits a substantial deviation relative to the preceding time step $t-1$:
\begin{equation}
    \mathcal{H}_{sens} = \left\{ t \mid \left| \frac{y_t - y_{t-1}}{y_{t-1}} \right| > 0.5 \right\}
\end{equation}
We strictly report S@25 and SMAPE for this subset to quantify the model's capability in simulating dynamic clinical transitions ($\Delta > 50\%$).

\subsection{Label Event Metrics}
For discrete clinical outputs, such as microbiology culture results or qualitative nursing assessments, the output space is categorical. We evaluate the semantic correctness of the generated descriptions using standard classification metrics:
\begin{equation}
    \text{Precision} = \frac{TP}{TP + FP}, \quad \text{Recall} = \frac{TP}{TP + FN}
\end{equation}
\begin{equation}
    \text{F1} = 2 \cdot \frac{\text{Precision} \cdot \text{Recall}}{\text{Precision} + \text{Recall}}
\end{equation}
where $TP$, $FP$, and $FN$ denote True Positives, False Positives, and False Negatives, respectively.

\vspace{0.5em}

\section{Additional Analysis}
\subsection{Inference Efficiency}
\label{sec:efficiency}

\begin{figure}[t]
    \centering
    \includegraphics[width=\linewidth]{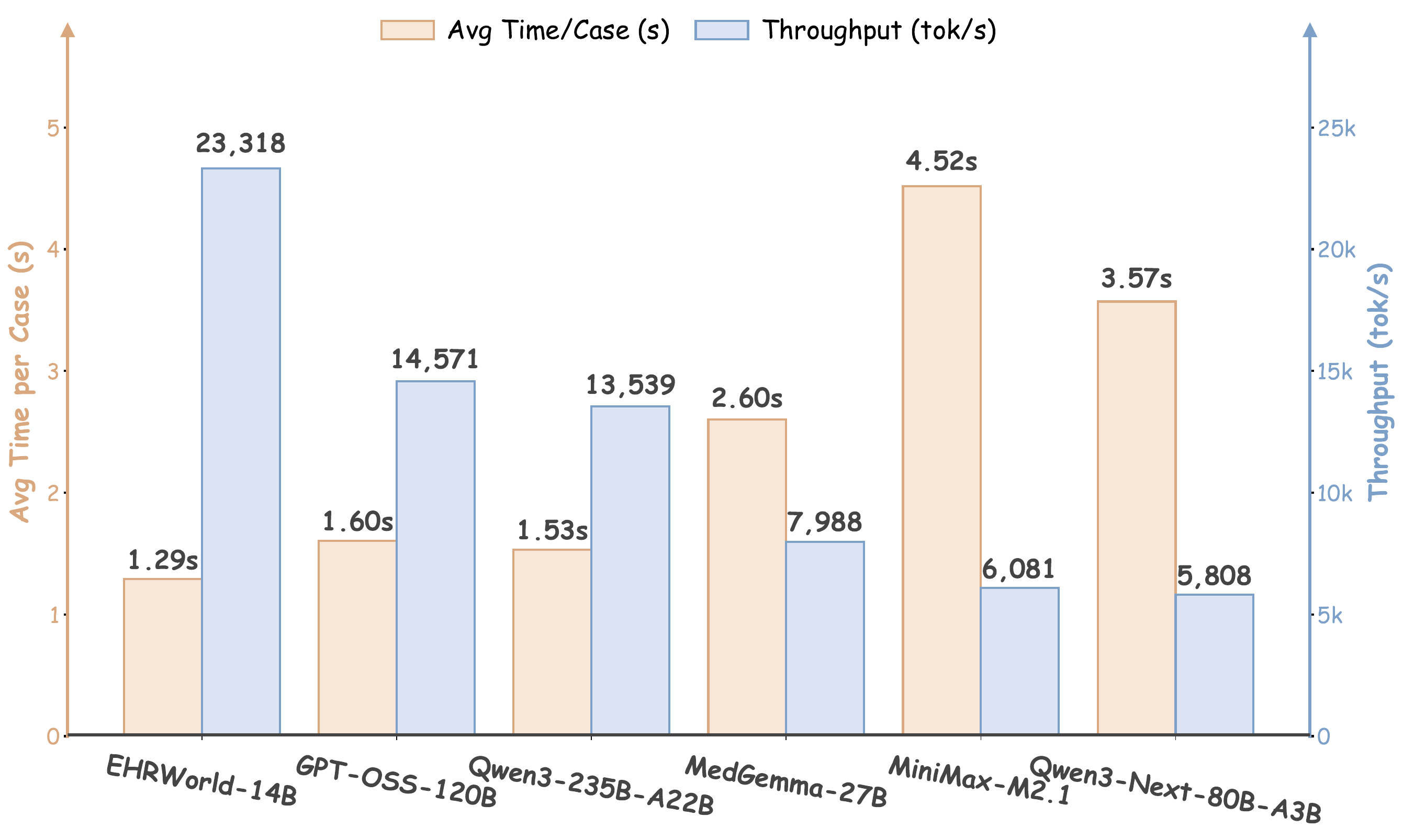}
    \caption{Comparison of inference time and token throughput across different models. \textit{EHRWorld-14B} demonstrates superior computational efficiency, achieving the fastest generation speed while maintaining the highest volume compared to all baselines.}
    \label{fig:efficiency}
\end{figure}

To assess deployment practicality for large-scale clinical simulation, we benchmark inference efficiency on a standardized cluster equipped with $8\times$ NVIDIA H100 GPUs. We restrict our comparison to competitive models that are viable for deployment on this specific hardware configuration. All models are served using the SGLang framework~\cite{zheng2024sglangefficientexecutionstructured} with a request concurrency set to 64. We report two key metrics: the average wall-clock time required to generate a full clinical case (Latency) and the aggregate system throughput (Tokens/s).

As illustrated in Figure~\ref{fig:efficiency}, \textbf{EHRWorld-14B} establishes a clear efficiency advantage across both metrics: \textbf{Latency reduction.}
Our model achieves the lowest average generation time of 1.29 seconds per case. This represents a significant speedup compared to domain-specific baselines such as \textit{MedGemma-27B-IT}, which requires 2.60 seconds per case—nearly double the inference time. Furthermore, \textit{EHRWorld-14B} maintains lower latency than massive general-purpose models, outperforming \textit{GPT-OSS-120B} (1.60s) and the MoE-based \textit{Qwen3-235B-A22B-Instruct} (1.53s), despite their highly optimized architectures.
\textbf{Throughput superiority.}
In terms of generation volume, \textit{EHRWorld-14B} reaches a peak throughput of 23,318 tokens/second. This is substantially higher than the runner-up, \textit{GPT-OSS-120B} (14,571 tokens/s), and offers an approximate $3\times$ to $4\times$ improvement over the heavier dense models like \textit{MiniMax-M2.1} (6,081 tokens/s) and \textit{Qwen3-Next-80B-A3B-Instruct} (5,808 tokens/s).

These results confirm that \textit{EHRWorld-14B} occupies a ``sweet spot'' in the trade-off between model scale and computational cost. By delivering high-fidelity clinical simulations with sub-1.5s latency and maximal throughput, it enables cost-effective, real-time applications that are computationally prohibitive for larger baseline models.

\subsection{Hyperparameter Sensitivity}
\label{app:ablation}

We examine the impact of the peak learning rate and training epochs on \textit{EHRWorld-14B} performance. The model shows sensitivity to these hyperparameters, as summarized below:

\begin{itemize}
    \item \textbf{Learning Rate:} $1 \times 10^{-6}$ achieves optimal performance. Higher rates cause instability, while lower rates result in slow convergence.
    \item \textbf{Epochs:} Performance plateaus at 3 epochs, with further training yielding minimal gains and increasing computational cost.
\end{itemize}

\begin{table}[h]
\centering
\small
\renewcommand{\arraystretch}{1.2}
\setlength{\tabcolsep}{6pt}
\caption{Ablation study on \textit{EHRWorld-14B}. Sensitivity to Learning Rate (LR) and Epochs is shown, with the final settings highlighted in \colorbox{oursbg}{color}.}
\label{tab:ablation}

\begin{tabular}{l c c c c}
\toprule
\textbf{Parameter} & \textbf{Value} & \textbf{S@25} & \textbf{Label F1} & \textbf{Avg Score} \\
\midrule
\rowcolor{groupbg} 
\multicolumn{5}{c}{\textit{\textbf{Learning Rate}} (Epochs fixed at 3)} \\
% LR 组开始
 & $1\times 10^{-7}$ & 0.652 & 0.885 & 0.720 \\
 & $5\times 10^{-7}$ & 0.698 & 0.902 & 0.751 \\
\rowcolor{oursbg}
\cellcolor{white} % 这一行确保 LR 所在的列保持白色，不被 oursbg 覆盖
 & $\mathbf{1\times 10^{-6}}$ & \textbf{0.716} & \textbf{0.913} & \textbf{0.765} \\
\multirow{-4}{*}{LR} % 注意：这里移到了最后一行，并使用了 -4
 & $5\times 10^{-6}$ & 0.645 & 0.860 & 0.702 \\
\midrule
\rowcolor{groupbg} 
\multicolumn{5}{c}{\textit{\textbf{Training Epochs}} (LR fixed at $1\times 10^{-6}$)} \\
% Epochs 组开始
 & 1 & 0.640 & 0.856 & 0.705 \\
 & 2 & 0.692 & 0.898 & 0.748 \\
\rowcolor{oursbg}
\cellcolor{white} % 同上，保持第一列背景为白
 & \textbf{3} & \textbf{0.716} & \textbf{0.913} & \textbf{0.765} \\
\multirow{-4}{*}{Epochs} % 移到了最后一行，使用 -4
 & 5 & 0.712 & 0.911 & 0.761 \\
\bottomrule
\end{tabular}
\end{table}

\end{document}